\documentclass[11pt]{article}
\usepackage{graphicx}
\usepackage{url}
\usepackage{amsmath}
\usepackage{booktabs}
\usepackage{hyperref}
\usepackage{caption}
\usepackage{xspace}
\usepackage{siunitx}
\usepackage{float}

\title{Efficient Hate Speech Detection: A Three-Layer LoRA-Tuned BERTweet Framework}

\author{Mahmoud El-Bahnasawi \\ Zewail City of Science and Technology \\ \texttt{s-mahmoud.elbahnasawi@zewailcity.edu.eg}}

\date{2025}

\begin{document}
\maketitle

\begin{abstract}
This paper addresses the critical challenge of developing computationally efficient hate speech detection systems that maintain competitive performance while being practical for real-time deployment. We propose a novel three-layer framework that combines rule-based pre-filtering with a parameter-efficient LoRA-tuned BERTweet model and continuous learning capabilities. Our approach achieves 0.85 macro F1 score - representing 94\% of the performance of state-of-the-art large language models like SafePhi (Phi-4 based) while using a base model that is \textbf{100$\times$ smaller} (134M vs 14B parameters). Compared to traditional BERT-based approaches with similar computational requirements, our method demonstrates superior performance through strategic dataset unification and optimized fine-tuning. The system requires only \textbf{1.87M trainable parameters} (1.37\% of full fine-tuning) and trains in approximately 2 hours on a single T4 GPU, making robust hate speech detection accessible in resource-constrained environments while maintaining competitive accuracy for real-world deployment.
\end{abstract}

\section{Introduction}
The rapid growth of harmful content on social platforms has created a persistent demand for hate speech detectors that are both accurate and deployable. Large language models (LLMs) achieve strong results, but their computational footprint makes them unsuitable for latency- and cost-sensitive settings. This paper asks: how can we approach LLM-level moderation quality while keeping the model small enough for real-time use?

We address this efficiency–performance gap with a three-layer framework built around a LoRA-tuned BERTweet encoder. At the high end of current approaches, systems such as SafePhi [1] report macro-F1 scores around 0.89 but rely on 14B-parameter foundations. At the low end, classic BERT or LSTM pipelines are affordable but plateau in accuracy. Our framework targets the middle: it keeps the domain-tuned advantages of BERTweet while restricting trainable parameters through PEFT.

Our contributions are threefold: (1) we show that LoRA applied to BERTweet can recover roughly 94\% of the performance reported for recent LLM-based moderators while using a 100$\times$ smaller base model; (2) we improve over prior efficient baselines through unified dataset construction and optimized fine-tuning; and (3) we present a production-oriented, three-layer architecture that short-circuits easy cases via rules and supports continuous adaptation.

\section{Related Work}

\subsection{Prior Surveys and Benchmarks}
The field of automated hate speech detection is characterized by rapid growth and significant methodological challenges. Recent systematic surveys have been instrumental in mapping this complex landscape. Baumler et al. \cite{baumler2025survey} conducted a comprehensive analysis of 46 publications, revealing critical field-wide issues: data scarcity (with 14 of 28 datasets containing only 5,000-15,000 samples), a heavy reliance on Twitter data leading to platform-specific biases, and a profound lack of methodological alignment. Their survey highlighted "an incoherent use of evaluation metrics" \cite{baumler2025survey} that prevents meaningful model comparison and identified a significant gap in the application of parameter-efficient fine-tuning techniques, like LoRA \cite{hu2021lora}, to hate speech detection. Furthermore, they noted that no existing work on multi-label hate speech classification utilized large foundation models, indicating an underexplored research direction.

Complementing this methodological overview, Tonneau et al. \cite{tonneau-etal-2024-languages} provided a crucial analysis of data biases, moving beyond language to examine geo-cultural representation. Their work found that English datasets drastically overrepresent US/UK contexts (73\%) while severely underrepresenting Global South English speakers from India (2\%), Nigeria (1\%), and Pakistan (0.3\%). This cultural bias poses a fundamental challenge to building robust and fair detection systems that perform equitably across diverse user bases.

\subsection{Existing Approaches and Models}
Current hate speech detection approaches can be broadly categorized into three paradigms based on their computational requirements and performance characteristics.

\textbf{Computationally Intensive LLM Approaches:} The most recent trend involves fine-tuning massive language models for hate speech detection. Machlovi et al. \cite{machlovi2025saferaimoderationevaluating} introduced the SafePhi model, a fine-tuned version of Phi-4--a 14-billion parameter small language model--which achieved a high macro-F1 of 0.89 on their unified HateBase benchmark. While demonstrating state-of-the-art performance, this approach requires full fine-tuning or adaptation of  large language models, creating prohibitive computational barriers for real-world deployment and iteration.

\textbf{Traditional Efficient Approaches:} At the other end of the spectrum, traditional methods prioritize efficiency over performance. Early work focused on logistic regression and SVM classifiers with hand-crafted features, while more recent approaches employ LSTMs and standard BERT variants. The Cardiff Unified model \cite{antypas2023robust} demonstrated that training on a combined corpus of 13 datasets (83K samples) yielded a model with superior cross-dataset generalization, achieving an average macro-F1 of 70.7\%. Similarly, MetaHate \cite{Piot_Martín-Rodilla_Parapar_2024} created a larger unified collection, aggregating 36 datasets into 1.2 million samples, with their BERT model achieving 0.89 accuracy and 0.80 macro-F1 using 110 million trainable parameters. While more efficient than LLM approaches, these methods face a performance ceiling that limits their practical effectiveness.

\textbf{Domain-Specific Foundation Models:} A key development in balancing efficiency and performance was BERTweet \cite{nguyen2020bertweet}, the first public large-scale language model pre-trained specifically on English tweets. With its training on 850 million tweets and 136 million parameters, BERTweet captures the informal, noisy, and idiosyncratic nature of Twitter language, establishing new state-of-the-art performance on various tweet NLP tasks. Its domain-specific advantages make it particularly suitable for hate speech detection on social media platforms while maintaining reasonable computational requirements.

These studies collectively highlight a persistent trade-off in the literature: between model performance and computational efficiency. \textbf{While SafePhi and similar LLM approaches demonstrate strong performance, they remain computationally prohibitive for real-time deployment. Conversely, traditional BERT and LSTM approaches offer reasonable efficiency but suffer from performance limitations. A significant gap remains in applying parameter-efficient fine-tuning (PEFT) methods to create high-performance, yet computationally lean, hate speech detection systems suitable for real-world deployment.} This research void is particularly critical given the documented challenges of data bias and the practical constraints of moderation systems that must operate at scale with limited computational resources.

\subsection{Hate Speech Datasets}
The foundation of any detection system is its data. The hate speech research community has developed a wide array of datasets, yet they are often limited in scale, scope, or accessibility \cite{baumler2025survey}. As previously noted, unifying these resources has been a key strategy to overcome these limitations.

For our work, we selected two unified datasets that align with the goal of robust model training. The first is the \textbf{Unified Human-Curated Moderation Dataset (HateBase)} \cite{machlovi2025saferaimoderationevaluating}, which comprises 236,738 instances across 49 fine-grained categories, advocating for a human-first approach to moderation. The second key resource is the \textbf{English Hate Speech Superset} \cite{tonneau-etal-2024-languages}, a comprehensive collection of approximately 360,000 samples that provides extensive geographic coverage while explicitly documenting its cultural biases. These datasets were chosen for their complementary strengths: HateBase offers fine-grained categorization and human-curated quality, while the English Hate Speech Superset provides substantial scale and well-documented geographic diversity. By building upon these consolidated benchmarks, our work aims to advance the state-of-the-art not by creating yet another dataset, but by demonstrating how to achieve superior efficiency and performance on these existing, community-relevant benchmarks.

\section{Methodology}

\subsection{Three-Layer Architecture}
\begin{figure}[H]
  \centering
  \includegraphics[width=0.85\linewidth]{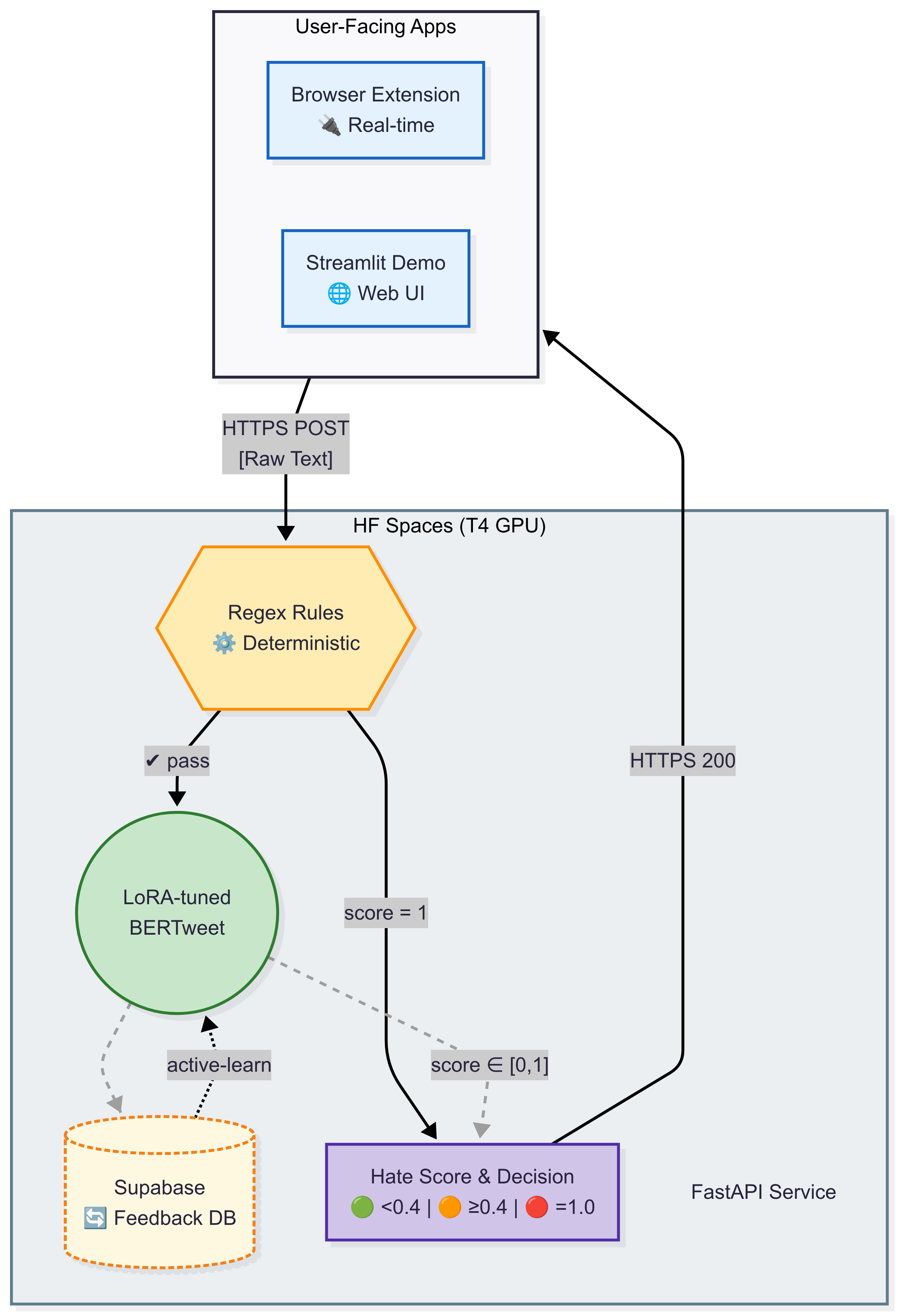}
  \caption{Three-layer moderation system architecture showing rule-based filtering (Layer 1), AI detection (Layer 2), and continuous learning (Layer 3).}
  \label{fig:system_architecture}
\end{figure}

Our framework implements a comprehensive three-layer moderation pipeline designed for both efficiency and accuracy in real-world deployment scenarios. The complete system architecture is illustrated in Figure~\ref{fig:system_architecture} (Appendix A), and operates as follows:

\textbf{Layer 1: Rule-based Pre-filtering} employs curated lexicons and regular expressions for instantaneous detection of explicit hate speech patterns. This layer utilizes comprehensive keyword databases including explicit profanity, hate speech terminology, extremist hashtags, and coded hate expressions. By handling clear-cut cases deterministically, this layer reduces computational load by avoiding model inference for a substantial portion of content in production deployment, significantly improving system throughput and reducing latency for real-time applications.

\textbf{Layer 2: AI-Powered Detection} utilizes LoRA-tuned BERTweet for nuanced hate speech identification, processing content that evades simple pattern matching. This layer provides contextual understanding and handles subtle or emerging hate speech patterns that cannot be captured by static rules alone.

\textbf{Layer 3: Continuous Learning} incorporates user feedback through a Supabase database for periodic model refinement. While the architecture is fully implemented with feedback collection and storage capabilities, no retraining has been performed using this data in the current evaluation. This layer is designed to enable future adaptation to emerging hate patterns while maintaining model stability through LoRA-only fine-tuning.

\begin{table}[H]
  \centering
  \caption{Decision Logic Across Moderation Layers}
  \begin{tabular}{lccc}
  \toprule
  Hate Score & Action & Moderation Layer & Rationale \\
  \midrule
  $ < 0.40 $ & Allow & -- & Below detection threshold \\
  0.40 - 0.99 & Block & AI Detection & Contextual hate speech \\
  = 1.00 & Block & Rule-based & Explicit hate content \\
  \bottomrule
  \end{tabular}
  \label{tab:decision_logic}
\end{table}

\subsection{Dataset Preparation}
Our training corpus integrates two key resources: the Unified Human-Curated Moderation Dataset (HateBase) \cite{machlovi2025saferaimoderationevaluating} and the English Hate Speech Superset from Tonneau et al. \cite{tonneau-etal-2024-languages}. This strategic combination provides comprehensive coverage of hate speech patterns while maintaining data diversity. The initial datasets comprised approximately 590K samples (HateBase: 230K + English Hate Speech Superset: 360K), which after preprocessing and deduplication resulted in approximately 530K labeled tweets.

The data preprocessing pipeline included dataset integration combining HateBase with the English Hate Speech Superset, removal of 14.7K duplicate entries across the combined datasets, text normalization through lowercasing and removal of emojis/URLs/mentions, length filtering excluding texts exceeding 60 words (approximately 35K rows), and tokenization using the BERTweet tokenizer for domain-appropriate processing.

We employed stratified splitting to maintain a consistent 67\% non-hate vs. 33\% hate distribution across train (490K samples), validation (30K samples), and test (10K samples) sets. We explicitly acknowledge the geographic biases documented by Tonneau et al. \cite{tonneau-etal-2024-languages}, with our dataset predominantly representing Western English variants while underrepresenting Global South contexts.

\subsection{Model Fine-tuning}

\subsubsection{Base Model and LoRA Configuration}
We employ BERTweet-base \cite{nguyen2020bertweet} as our foundation model, selected for its domain-specific pre-training on 850M English tweets. The model's \textbf{134.9 million parameters} are distributed across key components as detailed in Table~\ref{tab:param_distribution}.

\begin{table}[H]
  \centering
  \caption{Parameter Distribution of the BERTweet-base Model}
  \begin{tabular}{lccc}
  \toprule
  Section & Parameters & \% of Total & Description \\
  \midrule
  Embeddings & 49,254,912 & 36.51\% & Token, position, and type embeddings \\
  Encoder & 85,054,464 & 63.05\% & 12-layer transformer stack \\
  Classifier & 592,130 & 0.44\% & Task-specific classification head \\
  \midrule
  \textbf{Total} & \textbf{134,901,506} & \textbf{100.00\%} & \\
  \bottomrule
  \end{tabular}
  \label{tab:param_distribution}
\end{table}

For parameter-efficient fine-tuning, we implement Low-Rank Adaptation (LoRA) \cite{hu2021lora} with the following configuration:

LoRA was applied to the linear projections within the self-attention mechanism, specifically the query, key, value, and attention output dense layers. Complete architectural diagrams showing both the system overview (Figure~\ref{fig:system_architecture}) and model structure (Figure~\ref{fig:model_architecture}) are provided in Appendix A.

The complete set of trainable parameters consisted of LoRA adapters in the Q, K, V, and output dense layers of the self-attention blocks with rank=16 and alpha=12, along with the classifier layer which was made fully trainable via the \texttt{modules\_to\_save} argument to ensure optimal task-specific adaptation.

This configuration resulted in training only \textbf{1.87M parameters} out of 134.9M total (1.37\%), representing a 78× reduction in trainable parameters compared to full fine-tuning while maintaining competitive performance.

\subsubsection{Optimization and Training Strategy}
Our training regimen employs several optimization techniques to maximize efficiency on resource-constrained hardware. We use cosine decay learning rate scheduling with 10\% warmup, selected for its smooth, non-linear decay profile that provides aggressive learning in early epochs followed by fine-tuned adjustments in later stages. Mixed precision training (FP16) halves memory requirements while maintaining numerical stability. Gradient accumulation with steps of 2 achieves an effective batch size of 512 to fit within T4 GPU memory constraints. Gradient checkpointing is enabled to trade compute for memory, allowing for larger effective batch sizes within the same GPU memory budget. We use the AdamW optimizer with fused implementation, learning rate of 2e-3, weight decay of 0.01, and gradient norm clipping at 1.0. The complete training required 3 epochs taking approximately 2 hours on a single T4 GPU.

\subsection{Metric Choice}
We selected the Matthews Correlation Coefficient (MCC) as our primary optimization metric due to its robustness for imbalanced classification tasks. While macro F1 remains the standard reporting metric in hate speech literature for cross-study comparison, MCC's comprehensive evaluation of all confusion matrix categories makes it superior for hyperparameter optimization on imbalanced datasets. Our final configuration achieved optimal performance-efficiency balance using MCC as the optimization criterion while maintaining competitive macro F1 scores.

\section{Experiments and Results}

\subsection{Experimental Setup}
We evaluate our framework on the unified corpus described in Section 3.2 (530K instances after preprocessing), maintaining the original 67\% non-hateful vs. 33\% hateful class distribution. All experiments were conducted on a single NVIDIA T4 GPU to demonstrate practical resource constraints.

\subsection{Performance Evaluation}
Table~\ref{tab:performance} compares our model against recent efficiency-aware hate speech systems. Our LoRA-tuned BERTweet achieves 0.85 macro F1 and 0.68 MCC, outperforming Cardiff Unified (2023) and approaching the performance of the much larger SafePhi system. Specifically, our model recovers roughly 94\% of SafePhi's macro F1 while using a base encoder that is two orders of magnitude smaller (134M vs. 14B parameters).

\begin{table}[htbp]
\centering
\caption{Performance metrics comparison on the unified 530K corpus.}
\begin{tabular}{lcccc}
\toprule
Model & Accuracy & Macro F1 & MCC & Class Balance \\
\midrule 
MetaHate BERT (2024) & 0.88 & 0.80 & 0.61 & 80\%--20\% \\
Cardiff Unified (2023) & 0.70 & 0.707 & -- & 67\%--33\% \\
SafePhi (Machlovi et al., 2025) & -- & 0.89 & -- & 57\%--43\% \\
\textbf{Ours} & \textbf{0.86} & \textbf{0.85} & \textbf{0.68} & \textbf{67\%--33\%} \\
\bottomrule
\end{tabular}
\label{tab:performance}
\end{table}

\subsection{Efficiency and Scalability Analysis}
As shown in Table~\ref{tab:efficiency_comparison}, our approach trains only 1.87M parameters while leveraging 530K training samples. This represents a 78× reduction in trainable parameters compared to full BERTweet fine-tuning, enabling practical training in approximately 2 hours on a single T4 GPU without distributed setups.

\begin{table}[htbp]
\centering
\caption{Computational efficiency comparison. Trainable parameters refer to the part updated during fine-tuning.}
\begin{tabular}{lccc}
\toprule
Model & Base Model Size & Trainable Params & Training Data \\
\midrule
Cardiff Unified (2023) & 136M & -- & 83K \\
MetaHate BERT (2024) & 110M & 110M & 1.2M \\
SafePhi (2025) & 14B & -- & 236K \\
\textbf{Ours} & \textbf{134M} & \textbf{1.87M} & \textbf{530K} \\
\bottomrule
\end{tabular}
\label{tab:efficiency_comparison}
\end{table}

\subsection{Architectural Efficiency and Trade-offs}
The three-layer architecture provides additional efficiency gains beyond parameter reduction. Layer 1's rule-based filtering shifts computational cost from GPU-bound inference to efficient pattern matching, improving overall system throughput. While Layer 3's continuous learning capability is architecturally implemented, its longitudinal benefits require future evaluation with real user feedback.

Overall, our experimental results demonstrate that competitive hate speech detection (within 94\% of state-of-the-art LLM performance) can be achieved without massive models, offering a favorable operating point on the accuracy-efficiency curve for latency- and cost-constrained environments.

\section{Limitations and Future Directions}

Several limitations merit consideration in evaluating our framework. First, while our three-layer architecture demonstrates strong performance on explicit hate speech patterns, it faces challenges in detecting \textbf{implicit hate speech} – content that uses coded language, sarcasm, or cultural context to convey harmful meaning without explicit keywords. This remains a fundamental challenge across hate speech detection systems.

Second, the \textbf{continuous learning component (Layer 3)} represents a conceptual framework rather than a fully implemented system in our current evaluation. While we've designed the architecture for periodic model refinement through user feedback stored in Supabase, the actual implementation and longitudinal evaluation of this adaptive learning capability require further development and testing.

Additionally, we acknowledge the persistent \textbf{geo-cultural biases} in our training data, which predominantly represent Western English variants while underrepresenting Global South contexts. This limitation, inherited from the source datasets, may affect the model's fairness and performance across diverse cultural and linguistic communities.

Future work will address these limitations through several directions: (1) Enhanced implicit hate speech detection through multi-modal approaches and contextual understanding; (2) Full implementation and evaluation of the continuous learning pipeline with real user feedback; (3) Culturally-adaptive fine-tuning strategies and multilingual expansion while maintaining parameter efficiency; (4) Exploration of explainable AI techniques to improve model transparency for human moderators.

\section{Conclusion}
This work shows that high-coverage hate speech moderation does not require full fine-tuning of foundation models. By leveraging a domain-specific encoder (BERTweet), applying LoRA adapters to attention projections, and embedding the model inside a three-layer pipeline, we achieve a macro-F1 of 0.85 on a 530K-sample unified corpus while training only 1.87M parameters. This yields performance close to recent LLM-based moderators but at a fraction of the computational cost, and within a training budget feasible on a single T4 GPU. The production-oriented design, with a fast rule-based front layer and a feedback-driven adaptation layer, targets realistic deployment environments where most content is non-abusive and inference cost matters. Future work should strengthen the implicit-hate and emoji-rich cases, broaden geo-cultural coverage beyond Western English, and operationalize the continuous-learning layer against real user feedback.

To support reproducibility and practical adoption, we release our trained model \cite{huggingfacemodel2025}, implementation code \cite{githubrepo2025}, and an interactive demo \cite{streamlitdemo2025}. These resources provide immediate access to our hate speech detection framework for both research and production use.

\section*{Acknowledgements}
The authors would like to express their sincere gratitude to Dr. Doaa Shawky for her invaluable guidance, continuous support, and insightful feedback throughout this research. Her expertise and dedication were instrumental in making this work possible.

This work was supported by Zewail City of Science and Technology.

\bibliographystyle{unsrt}
\bibliography{references}

\appendix
\section{System and Model Architecture}
\label{app:architectures}

\begin{figure}[H]
  \centering
  \includegraphics[width=0.85\linewidth]{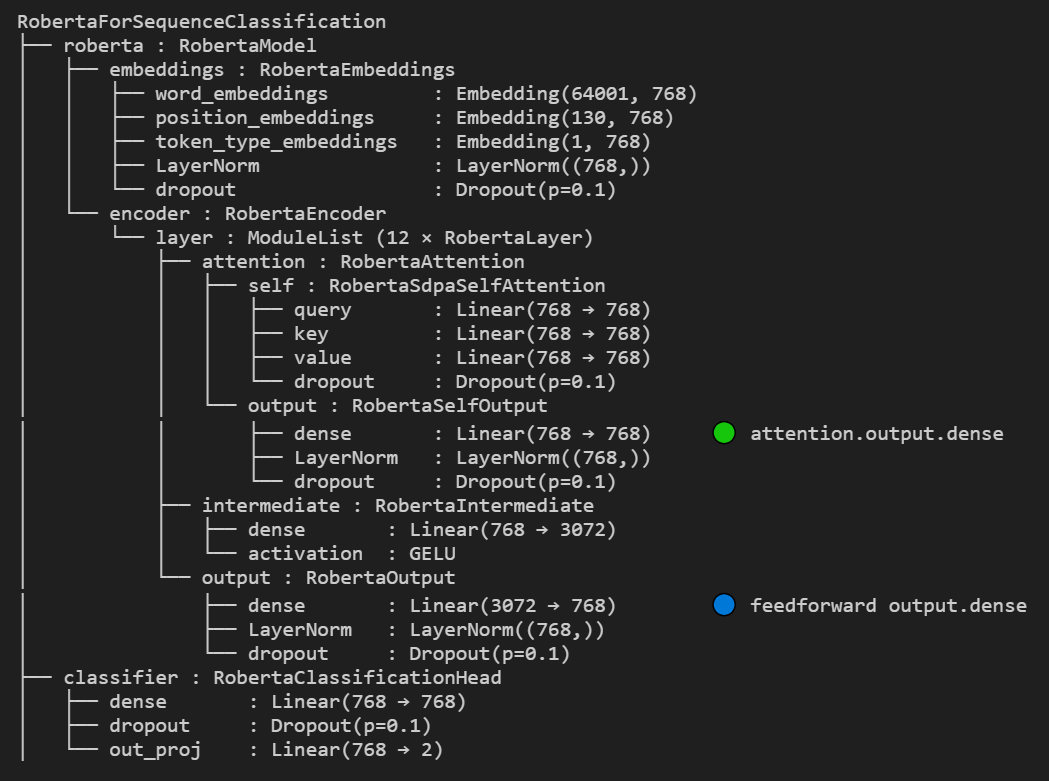}
  \caption{RobertaForSequenceClassification architecture (BERTweet-base) with LoRA adapters applied to attention layers (query, key, value, output.dense).}
  \label{fig:model_architecture}
\end{figure}

\end{document}